%% file: main.tex
\documentclass[10pt,twocolumn,letterpaper]{article}

\usepackage[pagenumbers]{cvpr} 

\usepackage{graphicx}
\usepackage{amsmath}
\usepackage{amssymb}
\usepackage{booktabs}
\usepackage{dsfont}
\usepackage{float}
\usepackage{xcolor}
\usepackage{wrapfig}
\usepackage{subfigure}
\usepackage[symbol]{footmisc}

\usepackage[pagebackref,breaklinks,colorlinks]{hyperref}
\usepackage{array}

\newcolumntype{L}[1]{>{\raggedright\let\newline\\\arraybackslash\hspace{0pt}}m{#1}}
\newcolumntype{C}[1]{>{\centering\let\newline\\\arraybackslash\hspace{0pt}}m{#1}}
\newcolumntype{R}[1]{>{\raggedleft\let\newline\\\arraybackslash\hspace{0pt}}m{#1}}

\usepackage[capitalize]{cleveref}
\crefname{section}{Sec.}{Secs.}
\Crefname{section}{Section}{Sections}
\Crefname{table}{Table}{Tables}
\crefname{table}{Tab.}{Tabs.}

\def\cvprPaperID{5115} 
\def\confName{CVPR}
\def\confYear{2022}

\colorlet{vancolor}{red!33!blue!66!}

\begin{document}


\title{PoseKernelLifter: Metric Lifting of 3D Human Pose using Sound}
\author{Zhijian Yang$^{1,2 *}$,
Xiaoran Fan$^{1}$, Volkan Isler$^{1,3 *}$,
Hyun Soo Park$^{1,3 *}$\\ 
$^{1}$Samsung AI Center NY, $^{2}$University of Illinois Urbana Champaign,\\ $^{3}$University of Minnesota Twin Cities
\\
\small saicny@samsung.com
}


\twocolumn[{%
\maketitle
	\begin{center}
		\centering
		\vspace{-5mm}
	 \includegraphics[width=1\linewidth]{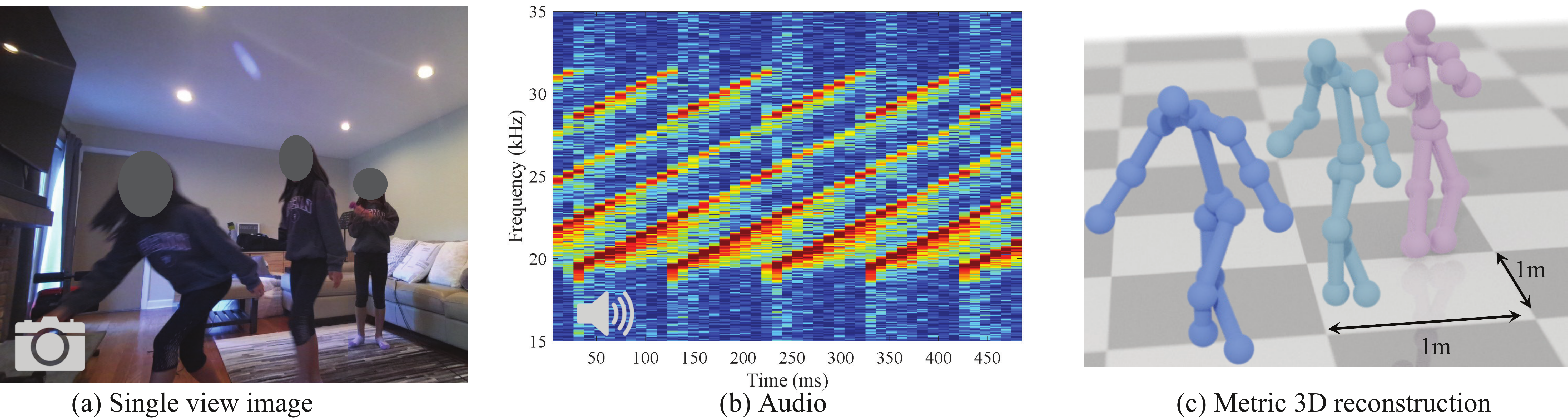}
	 \vspace{-6mm}
	\captionof{figure}{\small We present a new method for metric reconstruction of a human's pose from a single image along with audio signals transmitted from consumer-grade speakers. Our method recovers the metric scale by leveraging the fact that sound travels at a constant speed through a fixed medium. } 
\label{fig:teaser}
\end{center}	
}]
\footnotetext{* Work performed as a member of Samsung AI Center NY.}
\begin{abstract}
\input{abs}
\end{abstract}


\input{intro}
\input{related_work}
\input{method}

\input{dataset_eval}

\input{result}
\input{limitations}

\noindent\textbf{Disclosure} While VI and HSP are employees of the University of Minnesota, this work was solely done at Samsung AI Center New York.
{\small
\bibliographystyle{ieee_fullname.bst}
\bibliography{ref}
}

\end{document}

%% file: abs.tex
Reconstructing the 3D pose of a person in metric scale from a single view image is a geometrically ill-posed problem. For example, we can not measure the exact distance of a person to the camera from a single view image without additional scene assumptions (e.g., known height). Existing learning based approaches circumvent this issue by reconstructing the 3D pose up to scale. However, there are many applications such as virtual telepresence, robotics, and augmented reality that require metric scale reconstruction. In this paper, we show that audio signals recorded along with an image, provide complementary information to reconstruct the metric 3D pose of the person.
The key insight is that as the audio signals traverse across the 3D space, their interactions with the body provide metric information about the body's pose. 
Based on this insight, we introduce a time-invariant transfer function called pose kernel---the impulse response of audio signals induced by the body pose. The main properties of the pose kernel are that (1) its envelope highly correlates with 3D pose, (2) the time response corresponds to arrival time, indicating the metric distance to the microphone, and (3) it is invariant to changes in the scene geometry configurations. Therefore, it is readily generalizable to unseen scenes. We design a multi-stage 3D CNN that fuses audio and visual signals and learns to reconstruct 3D pose in a metric scale. We show that our multi-modal method produces accurate metric reconstruction in realworld scenes, which is not possible with state-of-the-art lifting approaches including parametric mesh regression and depth regression.

%% file: intro.tex
\section{Introduction}
Since the projection of the 3D world onto an image loses scale information, 3D reconstruction of a human's pose from a single image is an ill-posed problem.
To address this limitation, human pose priors have been used in existing lifting approaches~\cite{tome,habibie,chang,pavlakos,yushuke,llopart,li21} to reconstruct the plausible 3D pose given the 2D detected pose by predicting relative depths. The resulting reconstruction, nonetheless, still lacks \textit{metric scale}, i.e., the metric scale cannot be recovered without making an additional assumption such as known height or ground plane contact.
This fundamental limitation of 3D pose lifting precludes applying it to realworld downstream tasks, e.g., smart home facilitation, robotics, and augmented reality, where the precise metric measurements of human activities are critical, in relation to the surrounding physical objects.

In this paper, we study a problem of metric human pose reconstruction from a single view image by incorporating a new sensing modality---audio signals from consumer-grade speakers (Figure~\ref{fig:teaser}). Our insight is that while traversing a 3D environment, the transmitted audio signals undergo a characteristic transformation induced by the geometry of reflective physical objects including human body. This transformation is subtle yet highly indicative of the body pose geometry, which can be used to reason about the metric scale reconstruction. For instance, the same music playing in a room sounds differently based on the  presence or absence of a person, and more importantly, as the person moves.


We parametrize this transformation of audio signals using a time-invariant
transfer function called \textit{pose kernel}---an impulse response of audio induced by a body pose, i.e., the received audio signal is a temporal convolution of the transmitted signal with the pose kernel. Three key properties of pose kernel enables metric 3D pose lifting in a generalizable fashion: (1)~metric property: its impulse response is equivalent to the arrival time of the reflected audio, and therefore, it provides metric distance from the receiver (microphone); (2)~uniqueness: the envelope of pose kernel is strongly correlated with the location and pose of the target person; (3)~invariance: it is invariant to the geometry of surrounding environments, which allows us to generalize it to unseen environments.

While highly indicative of pose and location of the person in 3D, the pose kernel is a time-domain signal. Integrating it with the spatial-domain 2D pose detection is non-trivial. Further, generalization to new scenes requires precise 3D reasoning where existing audio-visual learning tasks such as source separation in an image domain and image representation learning~\cite{tian2021cyclic,ephrat:2018,gao:2019,owens2018audio} are not applicable. 


We address this challenge in 3D reasoning of visual and audio signals, by learning to fuse the pose kernels from multiple microphones and the 2D pose detected from an image, using a 3D convolutional neural network (3D CNN): (1) we project each point in 3D onto the image to encode the likelihood of landmarks (visual features); and (2) we spatially encode the time-domain pose kernel in 3D to form audio features. Inspired by the convolutional pose machine architecture~\cite{wei2016cpm}, a multi-stage 3D CNN is designed to predict the 3D heatmaps of the joints given the visual and audio features. This multi-stage design increases effective receptive field with a small convolutional kernel (e.g., $3\times 3\times 3$) while addressing the issue of vanishing gradients. 

In addition, we present a new dataset called \textit{PoseKernel} dataset. The dataset includes more than 
10,000 poses from six locations with more than six participants per location, 
performing diverse daily activities including sitting, drinking, walking, and jumping.
We use this dataset to evaluate the performance of our metric lifting method and show that it significantly outperforms state-of-the-art lifting approaches including mesh regression (e.g., FrankMocap~\cite{rong2021frankmocap}) and joint depth regression (e.g., Tome et al.~\cite{tome}). Due to the scale ambiguity of state-of-the-art approaches, the accuracy is dependent on the heights of target persons. In contrast, our approach can reliably recover the 3D poses regardless the heights, applicable to not only adults but also minors. 




\noindent\textbf{Why Metric Scale?} Smart home technology is poised to enter our daily activities, in particular, for monitoring fragile populations including children, patients, and the elderly. This requires not only 3D pose reconstruction but also holistic 3D understanding in the context of metric scenes, which allows AI and autonomous agents to respond in a situation-aware manner. While multiview cameras can provide metric reconstruction, the number of required cameras to cover the space increases quadratically as area increases. Our novel multi-modal solution can mitigate this challenge by leveraging multi-source audios (often inaudible) generated by consumer grade speakers (e.g., Alexa).

\noindent\textbf{Contributions} This paper makes a major conceptual contribution that sheds a new light on a single view pose estimation by incorporating with audio signals. The technical contributions include (1) a new formulation of the pose kernel that is a function of the body pose and location, which can be generalized to a new scene geometry, (2) the spatial encoding of pose kernel that facilitates fusing visual and audio features, (3) a multi-stage 3D CNN architecture that can effectively fuse them together, and (4) a strong performance of our method, outperforming state-of-the-art lifting approaches with meaningful margin. 



%% file: related_work.tex
\begin{figure*}[t].
\centering
\includegraphics[width=1\linewidth]{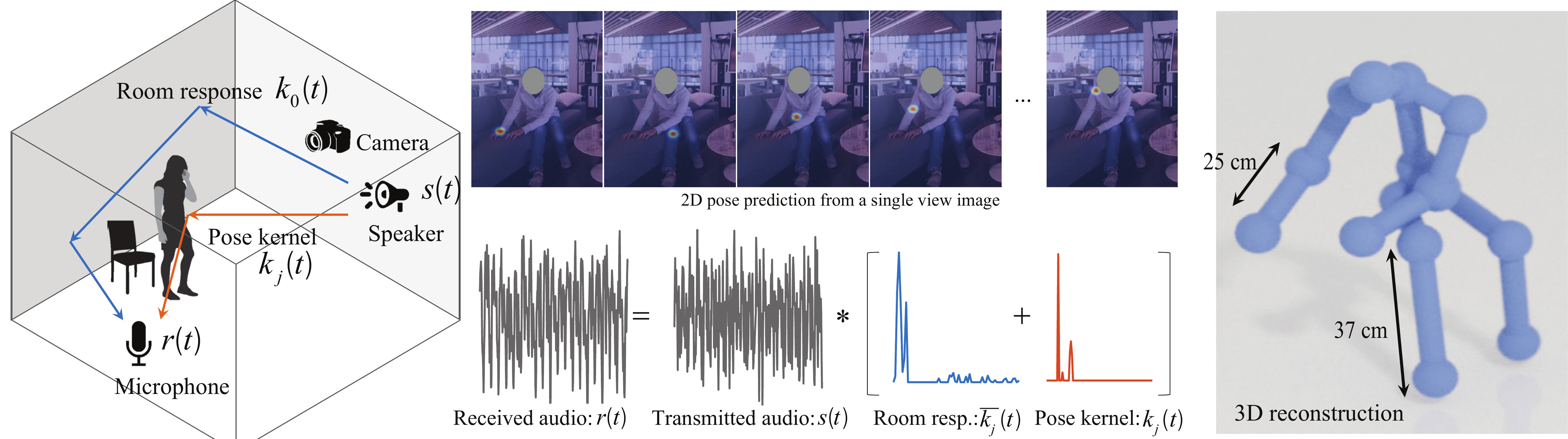}
\caption{(Left) Audio signals traverse in 3D across a room and are reflected by objects including human body surface. (Middle)~Given the received audio signals, we compute a human impulse response called \textit{pose kernel} by factoring out the room impulse response. (Right) We spatially encode the pose kernel in 3D space and combine it with the detected pose in an image using a 3D convolutional neural network to obtain the 3D metric reconstruction of human pose.}
\label{fig:pose_kernel} 
\vspace{-0.05in}
\end{figure*}

\section{Related work}

This paper is primarily concerned with integrating information from audio signals with single view 3D pose estimation to obtain metric scale. We briefly review the related work in these domains.

\noindent\textbf{Vision based Lifting} While reconstructing 3D pose (a set of body landmarks) from a 2D image is geometrically ill-posed, the spatial relationship between landmarks provides a geometric cue to reconstruct the 3D pose~\cite{cjtaylor}. This relationship can be learned from datasets that include 2D and 3D correspondences such as Human3.6M~\cite{ionescu2013human36m}, MPI-INF-3DHP~\cite{mono-3dhp2017} (multiview), Surreal~\cite{varol17_surreal} (synthetic), and 3DPW~\cite{vonMarcard2018} (external sensors). Given the 3D supervision, the spatial relationship can be directly learned via supervised learning~\cite{tome, sun2018,habibie,chang}. Various representations have been proposed to effectively encode the spatial relationship such as volumetric representation~\cite{pavlakos}, graph structure~\cite{cai19,zhao19,ci19,xu21}, transformer architecture~\cite{yushuke,llopart,li21}, compact designs for realtime reconstruction~\cite{vneck,xneck}, and inverse kinematics~\cite{li_cvpr21}. These supervised learning approaches that rely on the 3D ground truth supervision, however, show limited generalization to images of out-of-distribution scenes and poses due to the domain gap. Weakly supervised, self-supervised, and unsupervised learning have been used to address this challenge. For instance, human poses in videos are expected to move and deform continuously over time, leading to a temporal self-supervision~\cite{hossain}. A dilated convolution that increases temporal receptive fields is used to learn the temporal smoothness~\cite{pavllo,Tripathi20}, a global optimization is used to reconstruct temporally coherent pose and camera poses~\cite{arnab19}, and spatio-temporal graph convolution is used to capture pose and time dependency~\cite{cai19, yu20,liu20}. Multiview images provide a geometric constraint that allows learning view-invariant visual features to reconstruct 3D pose. The predicted 3D pose can be projected onto other view images~\cite{rhodin18, rhodin_eccv18,wendt}, stereo images are used to triangulate a 3D pose which can be used for 3D pseudo ground truth for other views~\cite{kocabas,iskakov, iqbal}, and epipolar geometry is used to learn 2D view invariant features for reconstruction~\cite{he20,yao19}. Adversarial learning enables decoupling the 3D poses and 2D images, i.e., 3D reconstruction from a 2D image must follow the distribution of 3D poses, which allows learning from diverse images (not necessarily videos or multiview)~\cite{chen,kudo,wandt19}. A characterization and differentiable augmentation of datasets, further, improves the generalization~\cite{gong, wang20}. With a few exceptions, despite remarkable performance, the reconstructed poses lack the metric scale because of the fundamental ambiguity of 3D pose estimation. Our approach leverages sound generated by consumer-grade speakers to lift the pose in 3D with physical scale.

\noindent\textbf{Multimodal Reconstruction}
Different modalities have been exploited for the purpose of 3D sensing and reconstruction, include RF based \cite{rfpose,rfavatar,wipose,jin2018towards,guan2020through}, inertial based \cite{shen2016smartwatch,yang2020ear}, and acoustic based \cite{yun2015turning,fan2021aurasense, fan2020acoustic,christensen2020batvision,wilson2021echo,senocak2018learning,chen2020soundspaces}.
Various applications including self-driving car \cite{guan2020through}, robot manipulation and grasping \cite{wang2016robot, wang2019multimodal, watkins2019multi, nadon2018multi}, simultaneous localization and mapping (SLAM) \cite{terblanche2021multimodal, doherty2019multimodal, akilan2020multimodality, singhal2016multi, sengupta2019dnn} benefited from multimodal reconstruction.
Audio, given its ambient nature, has attracted unique attention in multimodal machine learning \cite{liu2018towards,rodriguez2018methodology, ngiam2011multimodal,ghaleb2019metric, burnsmulti, mroueh2015deep}.
However, few works \cite{yun2015turning,christensen2020batvision,wilson2021echo} appear in the area of multimodal geometry understanding using audio as a modality, due to the heavy audio multipath posing various difficulties in 3D understanding.
Human pose, given its diverse nature, is especially challenging for traditional acoustic sensing, thus is sparsely studied.
While similar signals like WiFi and FMCW radio have been used for human pose estimation \cite{rfpose,rfavatar,wipose}, audio signal, given its lower speed of propagation, offers more accurate distance measurement than RF-based.

We address the challenge of audio multipath and uncover the potential of audio in accurate metric scale 3D human pose estimation. Specifically, we present the first method that combines audio signals with the 2D pose detection to reason about the 3D spatial relationship for metric reconstruction. Our approach is likely to be beneficial for various applications including smart home, AR/VR, robotics. 

%% file: method.tex
\section{Method}

We make use of audio signals as a new modality for metric human pose estimation. 
We learn a pose kernel that transforms audio signals, which can be encoded in 3D in conjunction with visual pose prediction as shown in Figure~\ref{fig:pose_kernel}.

\subsection{Pose Kernel Lifting}
We cast the problem of 3D pose lifting  as learning a function $g_{\boldsymbol{\theta}}$ that predicts a set of 3D heatmaps $\{\mathbf{P}_i\}_{i=1}^N$ given an input image $\mathbf{I} \in [0,1]^{W\times H \times 3}$ where 
$\mathbf{P}_i: \mathds{R}^3\rightarrow [0,1]$ is the likelihood of the $i^{\rm th}$ landmark over a 3D space, $W$ and $H$ are the width and height of the image, respectively, and $N$ is the number of landmarks. In other words, 
\begin{align}
\{\mathbf{P}_i\}_{i=1}^N = g_{\boldsymbol{\theta}} (\mathbf{I}), \label{Eq:pose}
\end{align}
where $g_{\boldsymbol{\theta}}$ a learnable function parametrized by its weights $\boldsymbol{\theta}$ that lift a 2D image to the 3D pose. Given the predicted 3D heatmaps, the optimal 3D pose is given by $\mathbf{X}^*_i = \underset{\mathbf{X}}{\operatorname{argmax}}~\mathbf{P}_i(\mathbf{X})$ so that $\mathbf{X}^*_i$ is the optimal location of the $i^{\rm th}$ landmark. In practice, we use a regular voxel grid to represent $\mathbf{P}$. 

We extend Equation~(\ref{Eq:pose}) by leveraging audio signals to reconstruct a metric scale human pose, i.e.,
\begin{align}
\{\mathbf{P}_i\}_{i=1}^N = g_{\boldsymbol{\theta}} (\mathbf{I}, \{k_j(t)\}_{j=1}^M), \label{Eq:time}
\end{align}
where $k_j(t)$ is the \textit{pose kernel} heard from the $j^{\rm th}$ microphone---a time-invariant audio impulse response with respect to human pose geometry that transforms the transmitted audio signals, as shown in Figure~\ref{fig:pose_kernel}. $M$ denotes the number of received audio signals\footnote{The number of audio sources (speakers) does not need to match with the number of received audio signals (microphones).}.
The pose kernel transforms the transmitted waveform as follows:
\begin{align}
r_j(t) = s(t) * (\overline{k}_j(t) + k_j(t)),
\end{align}
where $*$ is the operation of time convolution, $s(t)$ is the transmitted source signal and $r_j(t)$ is the received signal at the location of the $j^{\rm th}$ microphone. $\overline{k}_j(t)$ is the empty room impulse response that accounts for transformation of the source signal due to the static scene geometry, e.g., wall and objects, in the absence of a person. $k_j(t)$ is the pose kernel measured at the $j^{\rm th}$ microphone location that accounts for signal transformation due to human pose.

The pose kernel can be obtained using the inverse Fourier transform, i.e.,
\begin{align}
k_j(t) = \mathcal{F}^{-1} \{K_j(f)\},~~~K_j(f) = \frac{R_j(f)}{S(f)} - \overline{K}_j(f),
\end{align}
where $\mathcal{F}^{-1}$ is the inverse Fourier transformation, and $R_j(f)$, $S(f)$, and $\overline{K}_j(f)$ are the frequency responses of $r(t)$, $s(t)$, and $\overline{k}_j(t)$, respectively, e.g., $R(f) = \mathcal{F}\{r(t)\}$.


Since the pose kernel is dominated by direct reflection from the body, it is agnostic to scene geometry
\footnote{
The residual after subtracting the room response still includes multi-path effects involving the body.  However, we observe that such effects are negligible in practice, and the pose kernel is dominated by the direct reflection from the body. Therefore, it is agnostic to scene geometry. See Section~\ref{sec:discussion} for a discussion on multi-path shadow effect.}. The scene geometry is factored out by the empty room impulse response $\overline{k}_j(t)$ and the source audios $s(t)$ are canceled by the received audios $r(t)$, which allows us to generalize the learned $g_{\boldsymbol{\theta}}$ to various scenes. 

\subsection{Spatial Encoding of Pose Kernel}
\begin{wrapfigure}{r}{0.4\linewidth}
\vspace{-9mm}
  \begin{center}
    \includegraphics[width=1\linewidth]{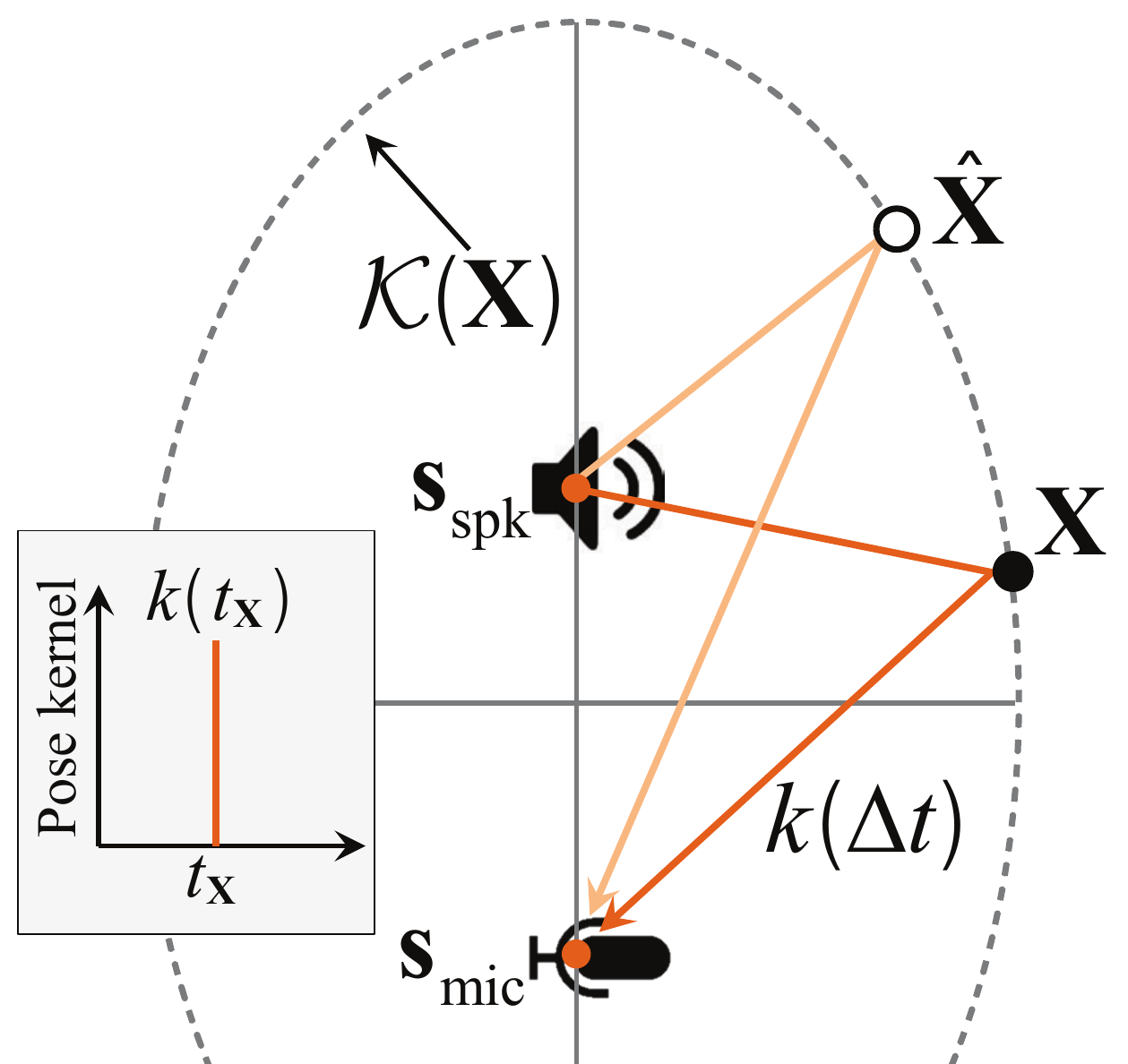}
  \end{center}
    \vspace{-5mm}
  \caption{Pose kernel spatial encoding.} 
  \label{Fig:spatial_encoding}
  \vspace{-5mm}
\end{wrapfigure}

We encode the time-domain pose kernel of the $j^{th}$ microphone, $k_j(t)$ to 3D spatial-domain where audio and visual signals can be fused. 
A transmitted audio at the speaker's location $\mathbf{s}_{\rm spk}\in \mathds{R}^3$ is reflected by the body surface at $\mathbf{X}\in \mathds{R}^3$ and arrives at the microphone's location $\mathbf{s}_{\rm mic}\in \mathds{R}^3$. The arrival time is:
\begin{align}
    t_{\mathbf{X}} = \frac{\|\mathbf{s}_{\rm spk} - \mathbf{X}\|+\|\mathbf{s}_{\rm mic} - \mathbf{X}\|}{v}, \label{Eq:delay}
\end{align}
where $t$ is the arrival time, and $v$ is the constant speed of sound (Figure~\ref{Fig:spatial_encoding}). 

The pose kernel is a superposition of impulse responses from the reflective points in the body surface, i.e.,
\begin{align}
    k_j(t) = \sum_{\mathbf{X}\in \mathcal{X}} A(\mathbf{X}) \delta(t-t_{\mathbf{X}}), \label{Eq:reflector}
\end{align}
where $\delta(t-t_{\mathbf{X}})$ is the Dirac delta function (impulse response) at $t= t_{\mathbf{X}}$. $t_{\mathbf{X}}$ is the arrival time of the audio signal reflected by the point $\mathbf{X}$ on the body surface $\mathcal{X}$. $A(\mathbf{X})$ is the reflection coefficient (gain) at $\mathbf{X}$.

Equation~(\ref{Eq:delay}) and (\ref{Eq:reflector}) imply two important spatial properties of the pose kernel. (i)
Since the locus of points whose sum of distances to the microphone and the speaker is an ellipsoid, Equation~(\ref{Eq:delay}) implies that the same impulse response can be generated by any point on this ellipsoid.
(ii) Due to the constant speed of sound, the response of the arrival time can be interpreted as that of the spatial distance by evaluating the pose kernel at the corresponding arrival time, $t_{\mathbf{X}}$:
\begin{align}
    \mathcal{K}_j(\mathbf{X}) = k_j(t)|_{t = t_{\mathbf{X}}},
    \label{Eq:encoding}
\end{align}
where $\mathcal{K}_j(\mathbf{X})$ is the spatial encoding of the pose kernel at $\mathbf{X}\in \mathds{R}^3$. 

\setlength{\columnsep}{10pt}
Let us illustrate the spatial encoding of pose kernel. Consider a point object $\mathbf{X} \in \mathds{R}^2$ that reflects an audio signal from the speaker $\mathbf{s}_{\rm spk}$ which is received by the microphone $\mathbf{s}_{\rm mic}$ as shown in Figure~\ref{Fig:spatial_encoding}. The received audio is delayed by $t_{\mathbf{X}}$, which can represented as a pose kernel $k(t) = A(\mathbf{X})\delta (t-t_{\mathbf{X}})$. This pose kernel can be spatially encoded as $\mathcal{K}(\mathbf{X})$ because the speed of the sound is constant. Note that there exists the infinite number of possible locations of $\mathbf{X}$ given the pose kernel because any point 
(e.g., $\widehat{\mathbf{X}}$) on the ellipse (dotted ellipse) has constant sum of distances from the speaker and microphone. 

\begin{figure}[t]
\begin{center}
\subfigure[Empty room response]{\label{fig:empty}\includegraphics[width=0.85\linewidth]{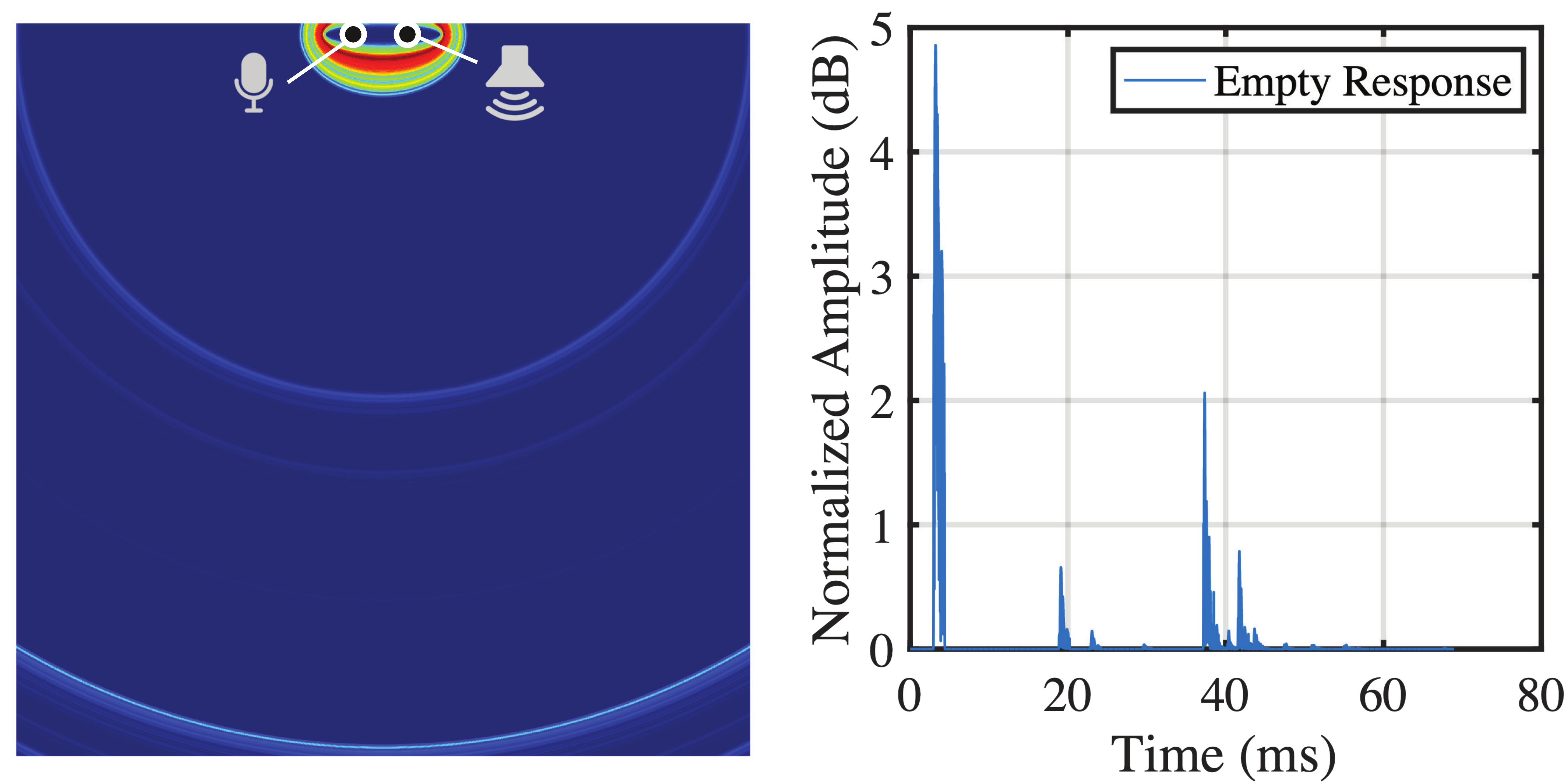}}\\\vspace{-3mm}
\subfigure[Object response]{\label{fig:object}\includegraphics[width=0.85\linewidth]{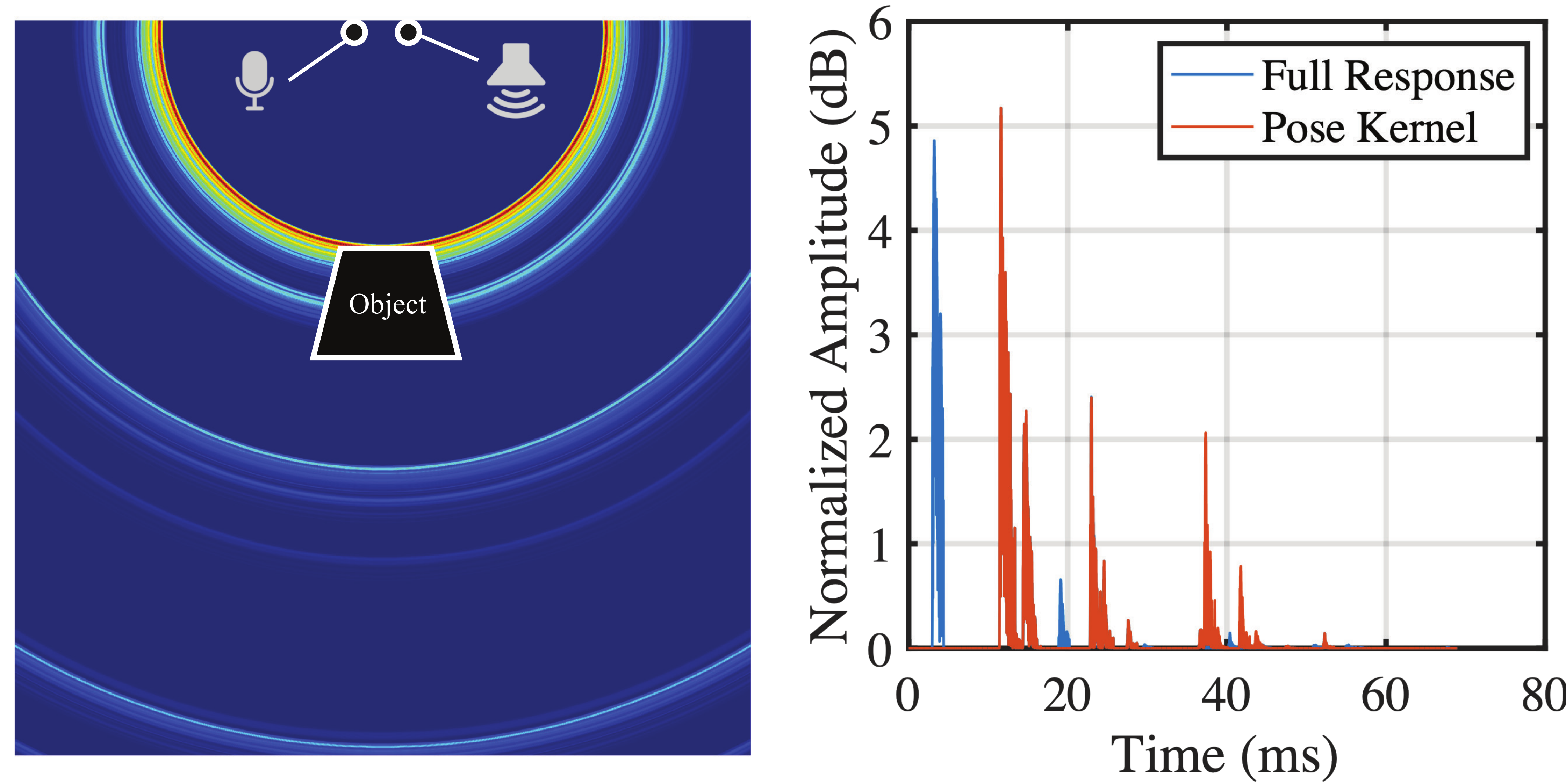}}\\\vspace{-3mm}
\subfigure[Rotated object response]{\label{fig:rotate}\includegraphics[width=0.85\linewidth]{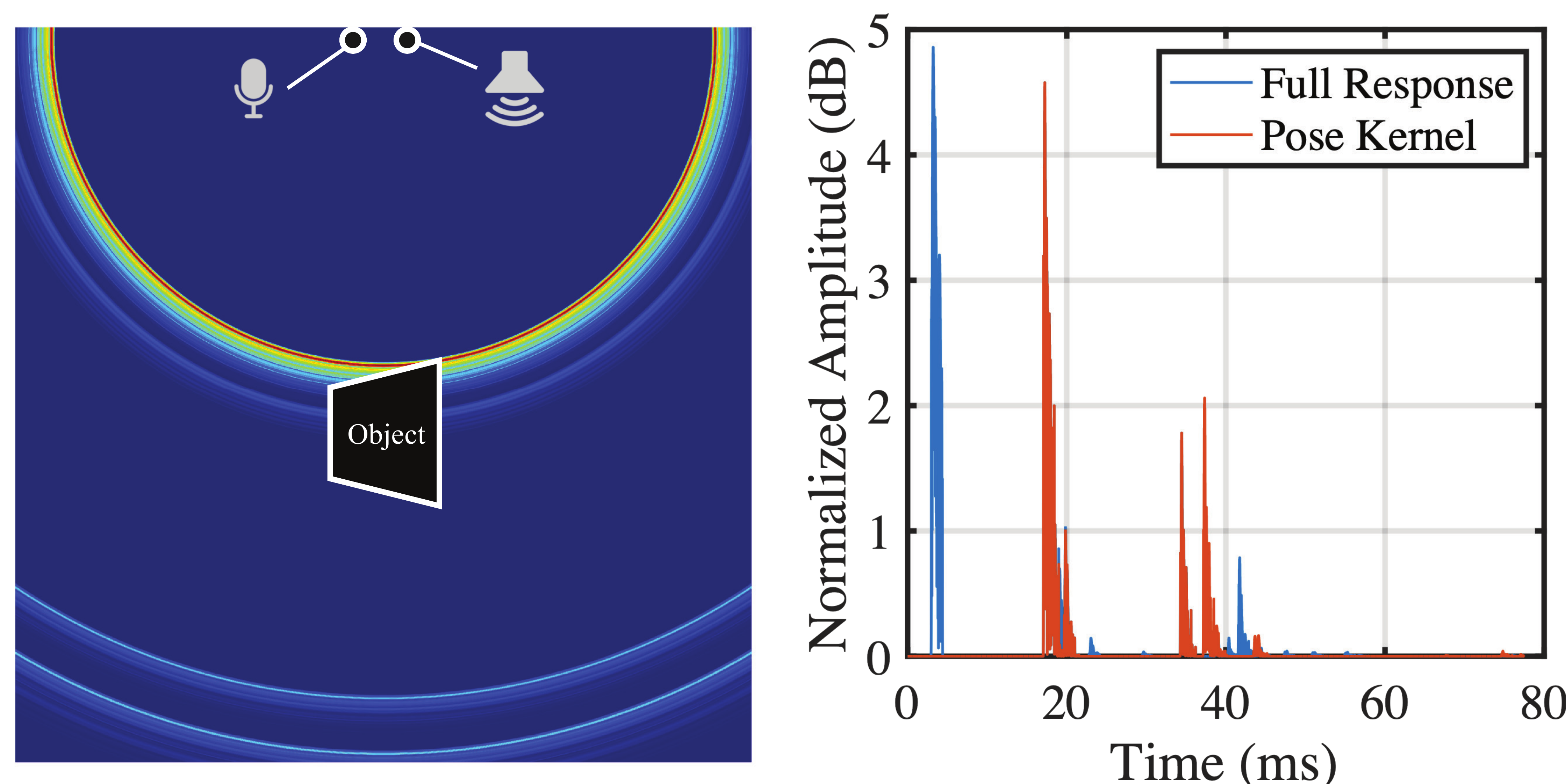}}\\\vspace{-3mm}
\subfigure[Translated object response]{\label{fig:translate}\includegraphics[width=0.85\linewidth]{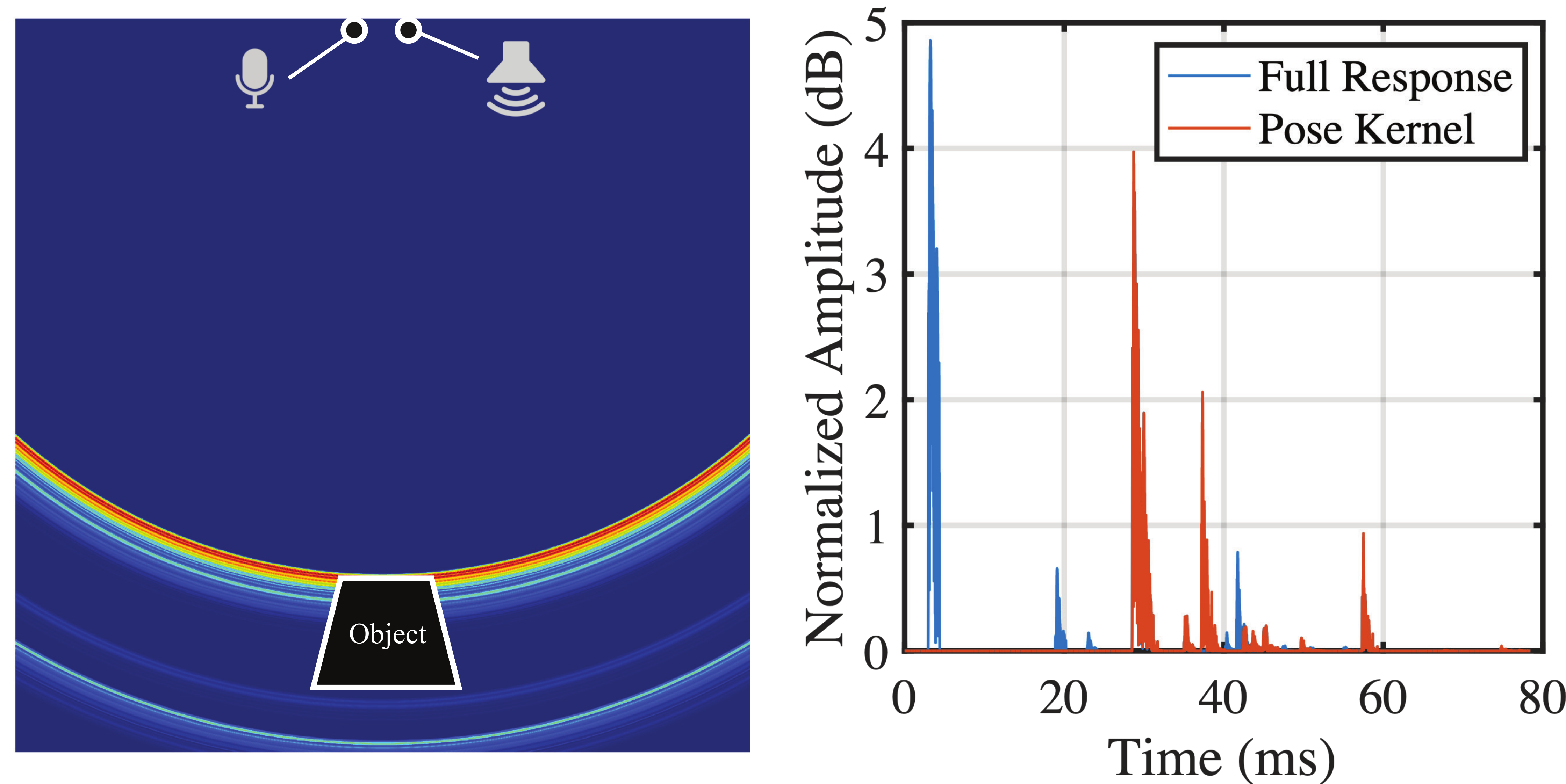}}
\end{center}
\vspace{-7mm}
  \caption{Visualization of the spatial encoding (left column) of time-domain impulse response (right column) through a sound simulation. The elliptical patterns can be observed by the spatial encoding where their focal points coincide with the locations of the speaker and microphone. (a) We visualize the empty room impulse response. (b) When an object is present, a strong impulse response that is reflected by the object surface can be observed. We show full responses that include the pose kernel. (b) Due to the object rotation, the kernel response is changed. (c) We observe delayed pose kernel due to translation. }
\label{fig:encoding_pose_kernel}
\vspace{-5mm}
\end{figure}

Figure \ref{fig:encoding_pose_kernel} illustrates (a) the empty room impulse response and (b,c,d) the full responses with the pose kernels by varying the location and pose of an object. 
The left column shows the pose kernel $k_j(t)$ encoded to the physical space, while the right column shows the actual signal.
Due to the fact that no bearing information is included from audio signal, each peak in the pose kernel $k_j(t)$ corresponds to a possible reflector location on the ellipse of which focal points coincide with the locations of the speaker and microphone.


With the spatial encoding of the pose kernel, we reformulate Equation~(\ref{Eq:time}):
\begin{align}
\{\mathbf{P}_i(\mathbf{X})\}_{i=1}^N = g_{\boldsymbol{\theta}} (\phi_v(\mathbf{X};\mathbf{I}),~~ \underset{j}{\operatorname{max}}~  \phi_a(\mathcal{K}_j(\mathbf{X}))), \label{Eq:space}
\end{align}
where $\phi_v$ and $\phi_a$ are the feature extractors for visual and audio signals, respectively. 

\begin{figure*}[t]
\centering
\includegraphics[width=\linewidth]{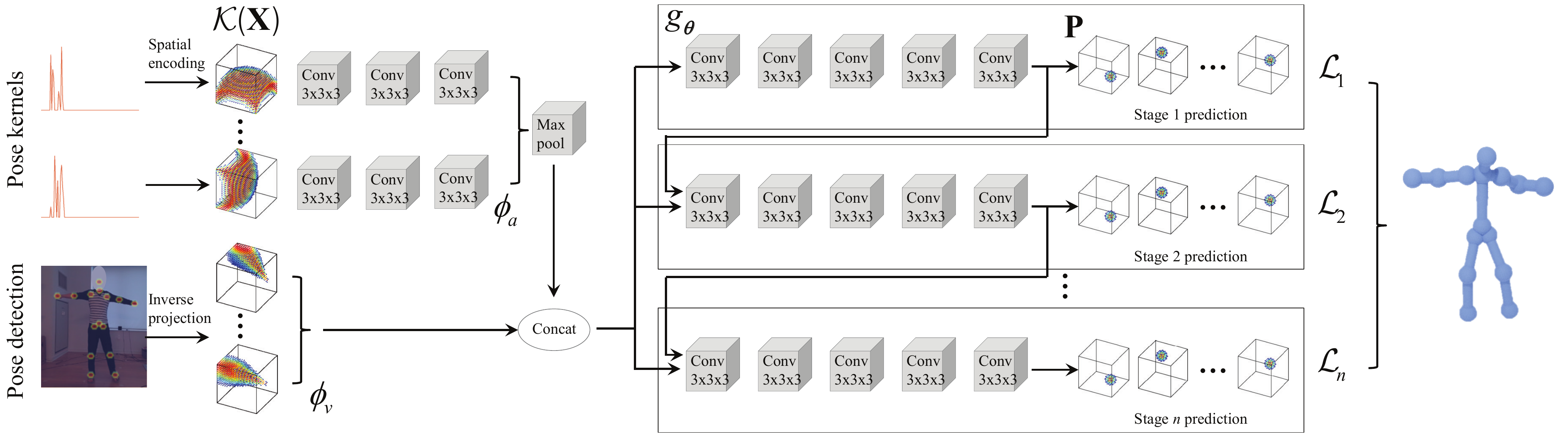}
\caption{We design a 3D convolutional neural network to encode pose kernels (audio) and 2D pose detection (image) to obtain the 3D metric reconstruction of a pose. We combine audio and visual features using a series of convolutions (audio features from multiple microphones are fused via max-pooling.). The audio visual features are convolved with a series of $3\times 3\times 3$ convolutional kernels to predict the set of 3D heatmaps for joints. We use multi-stage prediction, inspired by the convolutional pose machine architecture~\cite{wei2016cpm}, which can effectively increase the receptive field while avoiding vanishing gradients. }
\label{fig:architecture} 
\vspace{-0.05in}
\end{figure*}

Specifically, $\phi_v$ is the visual features evaluated at the projected location of $\mathbf{X}$ onto the image $\mathbf{I}$, i.e.,
\begin{align}
    \phi_v(\mathbf{X};\mathbf{I}) = \{\mathbf{p}_i(\Pi \mathbf{X})\}_{i=1}^N, 
\end{align}
where $\mathbf{p}_i \in [0,1]^{W\times H}$ is the likelihood of the $i^{\rm th}$ landmark in the image $\mathbf{I}$. $\Pi$ is the operation of 2D projection, i.e., $\mathbf{p}_i(\Pi \mathbf{X})$ is the likelihood of the $i^{\rm th}$ landmark at 2D projected location $\Pi \mathbf{X}$. 

$\phi_a(\mathcal{K}_j(\mathbf{X}))$ is the audio feature from the $j^{\rm th}$ pose kernel evaluated at $\mathbf{X}$. We use the max-pooling operation to fuse multiple received audio signals, which is agnostic to location and ordering of audio signals. 
This facilitates scene generalization where the learned audio features can be applied to a new scene with different audio configurations (e.g., the number of sources, locations, scene geometry).

We learn $g_{\boldsymbol{\theta}}$ and $\phi_a$ by minimizing the following loss:
\begin{align}
    \mathcal{L} = \sum_{\mathbf{I},\mathcal{K}, \widehat{\mathbf{P}}\in \mathcal{D}} \|g_{\boldsymbol{\theta}} (\phi_v,~ \underset{j}{\operatorname{max}}~  \phi_a(\mathcal{K}_j)) - \{\widehat{\mathbf{P}}_i\}_{i=1}^N\|^2,
\end{align}
where $\{\widehat{\mathbf{P}}_i\}_{i=1}^N$ is the ground truth 3D heatmaps, and $\mathcal{D}$ is the training dataset. Note that this paper focuses on the feasibility of metric lifting by using audio signals where we use an off-the-shelf human pose estimator $\{\mathbf{p}_i\}_{i=1}^N$~\cite{8765346}.

\subsection{Network Design and Implementation Details}

We design a 3D convolution neural network (3D CNN) to encode 2D pose detection from an image (using OpenPose~\cite{8765346}) and four audio signals from microphones. Inspired by the design of the convolution pose machine~\cite{wei2016cpm}, the network is composed of six stages that can increase the receptive field while avoiding the issue of the vanishing gradients. The 2D pose detection is represented by a set of heatmaps that are encoded in the $70\times 70\times 50$ voxel grid via inverse projection, which forms 16 channel 3D heatmaps. For the pose kernel from each microphone, we spatially encode over a $70\times 70\times 50$ voxel grid that are convolved with three 3D convolutional filters followed by max pooling across four audio channels. Each grid is 5 cm, resulting in 3.5 m$\times$3.5 m$\times$2.5 m space. These audio features are combined with the visual features to form the audio-visual features. These features are transformed by a set of 3D convolutions to predict the 3D heatmaps for each joint. The prediction, in turn, is combined with the audio-visual features to form the next stage prediction. 
The network architecture is shown in Figure \ref{fig:architecture}.

We implemented the network with PyTorch, and trained it on a server using 4 Telsa v100 GPUs. SGD optimizer is used, and learning rate is 1. The model has been trained for 70 epochs (around 36 hours) until convergence. 





%% file: dataset_eval.tex
\section{PoseKernel Dataset}

\begin{figure}[t]
\centering
\includegraphics[width=\linewidth]{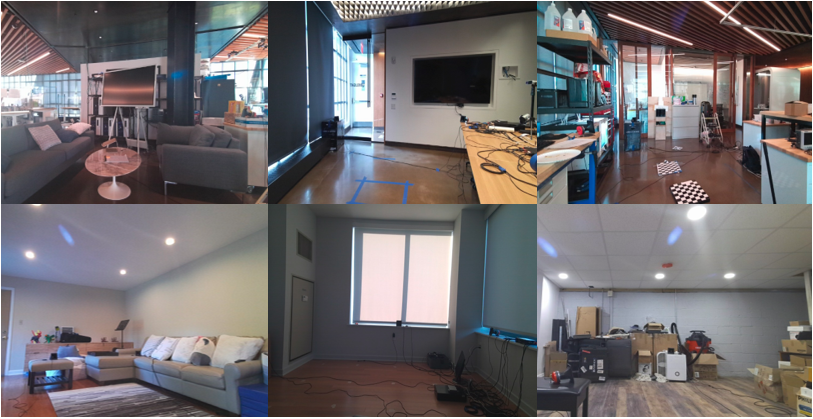}
\caption{We collect our PoseKernel dataset in different environments with at least six participants per location, totalling more than 10,000 poses. }
\label{fig:environments} 
\end{figure}

We collect a new dataset called \textit{PoseKernel} dataset. It is composed of more than 10,000 frames of synchronized videos and audios from six locations including living room, office, conference room, laboratory, etc. For each location, more than six participants were asked to perform as shown in Figure~\ref{fig:environments}. (We plan to release the dataset when paper publishes.)

The cameras, speakers, and microphones are spatially calibrated using off-the-shelf structure-from-motion software such as COLMAP~\cite{schoenberger2016sfm} by scanning the environments with an additional camera and use the metric depth from the RGB-D cameras to estimate the true scale of the 3D reconstruction. We manually synchronize the videos and speakers by a distinctive audio signal, e.g., clapping, and the speakers and microphones are hardware synchronized by a field recorder (e.g., Zoom F8n Recorder) at a sample rate of 96 kHz. 

For each scene, video data are captured by two RGB-D Azure Kinect cameras. These calibrated RGB-D cameras are used to estimate the ground truth 3D body pose using state-of-the-art pose estimation methods such as FrankMocap~\cite{rong2021frankmocap}. Multiple RGB-D videos are only used to generate the ground truth pose for training. In the testing phase, only a single RGB video is used. 

Four speakers and four microphones are used to generate the audio signals. Each speaker generates a chirp audio signal sweeping frequencies between 19 kHz to 32 kHz. We use this frequency band because it is audible from consumer-grade microphones while inaudible for humans. Therefore, it does not interfere with human generated audio. In order to send multiple audio signals from four speakers, we use frequency-division multiplexing within the frequency band. Each chirp duration is 100 ms, resulting in 10 FPS reconstruction. At the beginning of every capture session, we capture the empty room impulse response for each microphone in the absence of humans. 


We ask participants to perform a wide range of daily activities, e.g., sitting, standing, walking, and drinking, and range of motion in the environments. To evaluate the generalization on heights, in our test data, we include three minors (height between 140 cm and 150 cm) with consent from their guardians. All person identifiable information including faces are removed from the dataset.

%% file: result.tex
\begin{table*}[t]
\scriptsize
\begin{tabular}{p{2.1cm}|cccccccc|c}
\toprule 
Methods & Head & Neck & Hand & Elbow & Shoulder& Hip & Knee & Foot & Mean\\ 
\hline
\texttt{Vis.:LfD~\cite{tome}} & 57.49 / 34.77 & 52.31 / 29.89 & 59.09 / 45.01 & 57.10 / 39.30 & 55.22 / 32.77 & 56.31 / 32.81 & 50.76 / 51.09 & 54.59 / 57.89 & 55.76 / 41.43\\
\texttt{Vis.:Frank~\cite{rong2021frankmocap}} & 42.07 / 17.33 & 43.24 / 17.87 & 44.38 / \textbf{18.43} & 44.68 / 18.79 & 43.70 / 18.21 & 44.33 / 18.55 & 46.12 / 19.14 & 48.76 / 20.98 & 44.60 / 18.71 \\
\texttt{Audio$\times$4} & 313.4 / 290.0 & 321.2 / 259.1 & 253.6 / 277.9 & 303.6 / 265.0 & 143.5 / 143.8 & 350.0 / 288.7 & 240.2 / 261.0 & 85.3 / 145.4 & 255.1 / 229.5 \\
\texttt{Vis.+Audio$\times$2} &  10.13 / \textbf{12.45} & 10.02 / \textbf{10.22} & 12.69 / 21.64 & 12.55 / 15.77 & 11.40 / \textbf{11.47} & 12.44 / 9.70 & \textbf{13.84} / 13.81 & \textbf{15.19} / 16.29 & 12.23 / 13.46\\
\texttt{Ours}& \textbf{8.56} / 14.17 & \textbf{8.20} / 10.83 & \textbf{12.21} / 20.17 & \textbf{11.56} / \textbf{14.30} & \textbf{9.81} / 11.68 & \textbf{10.97} / \textbf{9.50} & 14.29 / \textbf{13.06} & 16.85 / \textbf{14.66} & \textbf{11.28} / \textbf{13.14}\\
\bottomrule 
\end{tabular}
\vspace{-3mm}
\caption{We use MPJPE (the lower, the better) as the evaluation metric to compare our method with state-of-the-art vision based algorithms including LfD (\texttt{Vis.:LfD}~\cite{tome}) and FrankMocap (\texttt{Vis.:Frank}~\cite{rong2021frankmocap}) and our ablated algorithms including \texttt{Audio$\times$4} and \texttt{Vision+Audio$\times$2}. We test on two sets: one with minor participants and one with adult participants (minor MPJPE/adult MPJPE). All numbers are reported in cm. }
\label{mpjpe}
\end{table*}

\begin{table*}[t]
\scriptsize
\begin{tabular}{p{2.3cm}p{1cm}|cccccccc|c}
\toprule 
Methods & $t$ & Head & Neck & Hand & Elbow & Shoulder &  Hip & Knee & Foot & Mean\\ 
\hline
\texttt{Vis.:LfD}~\cite{tome} & 10 cm &.000 / .068 & .008 / .152 & .000 / .051 & .000 / .055 & .000 / .131 & .000 / .063 & .008 / .017 & .000 / .017 & .001 / .064 \\
\texttt{Vis.:Frank}~\cite{rong2021frankmocap} & 10 cm & .033 / .253 & .025 / .257 & .016 / \textbf{.245} & .025 / .249 & .025 / .257 & .025 / .257 & .025 / .232 & .016 / .232 & .023 / .248 \\
\texttt{Audio$\times$4} &  10 cm & .000 / .000 & .000 / .000 & .000 / .000 & .000 / .000 & .000 / .000 & .000 / .000 & .000 / .000 & .000 / .000 & .000 / .000 \\
\texttt{Vis.+Audio$\times$2} & 10cm & .590 / \textbf{.409} & .541 / \textbf{.515} & \textbf{.410} / .224 & \textbf{.459} / .270 & \textbf{.557} / \textbf{.464} & .410 / \textbf{.578} & \textbf{.311} / .354 & \textbf{.238} / .325 & \textbf{.436} / .417 \\
\texttt{Ours} & 10 cm & \textbf{.648} / .380 & \textbf{.639} / .473 & .336 / .241 & .369 / \textbf{.367} & .484 / .439 & \textbf{.418} / .544 & .230 / \textbf{.397} & .213 / \textbf{.346} & .432 / \textbf{.417} \\
\hline 
\texttt{Vis.:LfD}~\cite{tome} & 20 cm & .000 / .405 & .033 / .586 & .000 / .262 & .000 / .350 & .016 / .468 & .008 / .498 & .041 / .122 & .000 / .097 & .011 / .320 \\
\texttt{Vis.:Frank}~\cite{rong2021frankmocap} &20 cm &.049 / .616 & .049 / .582 & .049 / \textbf{.586} & .049 / .557 & .049 / .582 & .049 / .561 & .049 / .544 & .049 / .489 & .049 / .568 \\
\texttt{Audio$\times$4} &  20 cm & .000 / .000 & .000 / .000 & .000 / .000 & .000 / .000 & .000 / .021 & .000 / .000 & .000 / .000 & .008 / .000 & .001 / .001 \\
\texttt{Vision+Audio$\times$2}&  20 cm & .861 / \textbf{.819} & .844 / \textbf{.899} & .820 / .540 & .820 / .713 & .811 / \textbf{.840} & .820 / .890 & .779 / .802 & \textbf{.811} / .738 & .815 / .794 \\
\texttt{Ours} & 20 cm & \textbf{.918} / .772 & \textbf{.943} / .844 & \textbf{.844} / .565 & \textbf{.885} / \textbf{.768} & \textbf{.918} / .814 & \textbf{.893} / \textbf{.911} & \textbf{.779} / \textbf{.831} & .639 / \textbf{.781} & \textbf{.861} / \textbf{.804} \\
\hline 
\texttt{Vis.:LfD}~\cite{tome} & 30 cm & .049 / .722 & .066 / .827 & .016 / .506 & .033 / .603 & .057 / .789 & .057 / .776 & .115 / .316 & .016 / .270 & .048 / .567 \\
\texttt{Vis.:Frank}~\cite{rong2021frankmocap} &30 cm & ..082 / .911 & .066 / .890 & .074 / .882 & .074 / .852 & .074 / .882 & .066 / .869 & .057 / .852 & .057 / .789 & .064 / .865 \\
\texttt{Audio$\times$4} &  30 cm & .000 / .000 & .000 / .000 & .000 / .000 & .000 / .000 & .000 / .025 & .000 / .000 & .000 / .000 & .033 / .004 & .002 / .003 \\
\texttt{Vis.+Audio$\times$2} & 30 cm & .967 / \textbf{.958} & .975 / .966 & .926 / .751 & .943 / .899 & .934 / .970 & .959 / .966 & .934 / .941 & .943 / .899 & .954 / .928 \\
\texttt{Ours} & 30 cm & \textbf{.992} / .941 & \textbf{1.000} / \textbf{.966} & \textbf{.967} / \textbf{.793} & \textbf{.959} / \textbf{.941} & \textbf{.992} / \textbf{.983} & \textbf{.992} / \textbf{.983} & \textbf{.959} / \textbf{.958} & \textbf{.959} / \textbf{.903} & \textbf{.980} / \textbf{.940} \\
\hline 
\texttt{Vis.:LfD}~\cite{tome} & 40 cm & .074 / .878 & .123 / .903 & .115 / .667 & .131 / .764 & .115 / .861 & .074 / .882 & .361 / .536 & .197 / .451 & .156 / .725 \\
\texttt{Vis.:Frank}~\cite{rong2021frankmocap} &40 cm & .418 / .987 & .336 / .979 & .303 / \textbf{.970} & .270 / .970 & .320 / .979 & .279 / .970 & .213 / .970 & .131 / .954 & .281 / .969 \\
\texttt{Audio$\times$4} &  40 cm & .000 / .000 & .000 / .000 & .000 / .000 & .000 / .000 & .000 / .055 & .000 / .000 & .000 / .000 & .074 / .013 & .005 / .007 \\
\texttt{Vis.+Audio$\times$2} &  40 cm & 1.000 / \textbf{.987} & 1.000 / .992 & .975 / .861 & .967 / .962 & .992 / .983 & .992 / .979 & .992 / .970 & .959 / .949 & .986 / .965 \\
\texttt{Ours} & 40 cm & \textbf{1.000} / .983 & \textbf{1.000} / \textbf{.996} & \textbf{.992} / .895 & \textbf{1.000} / \textbf{.983} & \textbf{1.000} / \textbf{.996} & \textbf{1.000} / \textbf{.996} & \textbf{1.000} / \textbf{.970} & \textbf{.984} / \textbf{.966} & \textbf{.997} / \textbf{.976} \\
\bottomrule 
\end{tabular}
\vspace{-3mm}
\caption{We use PCK@$t$ (the higher, the better) as the evaluation metric to compare our method with state-of-the-art vision based algorithms including LfD (\texttt{Vis.:LfD}~\cite{tome}) and FrankMocap (\texttt{Vis.:Frank}~\cite{rong2021frankmocap}) and our ablated algorithms including \texttt{Audio$\times$4}  and \texttt{Vis.+Audio$\times$2}. We test on two sets: one with minor participants and one with adult participants (minor PCK/adult PCK). }
\label{pck}
\end{table*}

\section{Results}
We evaluate our method on the PoseKernel dataset by comparing with state-of-the-art and baseline algorithms. 

\noindent\textbf{Evaluation Metric} We use the mean per joint position error (MPJPE) and the percentage of correct keypoint (PCK) in 3D as the main evaluation metrics. For PCK, we report the PCK\@$t$ where $t$ is the error tolerance in cm. 

\noindent\textbf{Baseline Algorithms} Two state-of-the-art baseline algorithms are used. (1) Lifting from the Deep, or \texttt{Vis.:LfD}~\cite{tome} is a vision based algorithm that regresses the 3D pose from a single view image by learning 2D and 3D joint locations 
together. To resolve the depth ambiguity, a statistical model is learned to  generate a plausible 3D reconstruction. This algorithm predicts 3D pose directly where we apply the Procrustes analysis to align with image projection.  (2) FrankMocap (\texttt{Vis.:FrankMocap}~\cite{rong2021frankmocap}) leverages the pseudo ground truth 3D poses on in-the-wild images that can be obtained by EFT~\cite{joo2020eft}. Augmenting 3D supervision improves the performance of 3D pose reconstruction. This algorithm predicts the shape and pose using the SMPL parameteric mesh model~\cite{SMPL:2015}. None of existing single view reconstruction approaches including these baseline methods produces metric scale reconstruction. 
Given their 3D reconstruction, we scale it to a metric scale by using the average human height in our dataset (1.7m). 

\noindent\textbf{Our Ablated Algorithms} In addition to the state-of-the-art vision based algorithms, we compare our method by ablating our sensing modalities. 
(1) \texttt{Audio$\times$4} uses four audio signals to reconstruct the 3D joint locations to study the impact of the 2D visual information. (2) \texttt{Vis.+Audio$\times$2} uses a single view image and two audio sources to predict the 3D joint location in the 3D voxel space. (3) \texttt{Ours} is equivalent to Vision+Audio$\times$4.



\subsection{PoseKernelLifter Evaluation}

\begin{figure*}[t]
\centering
\includegraphics[width=\linewidth]{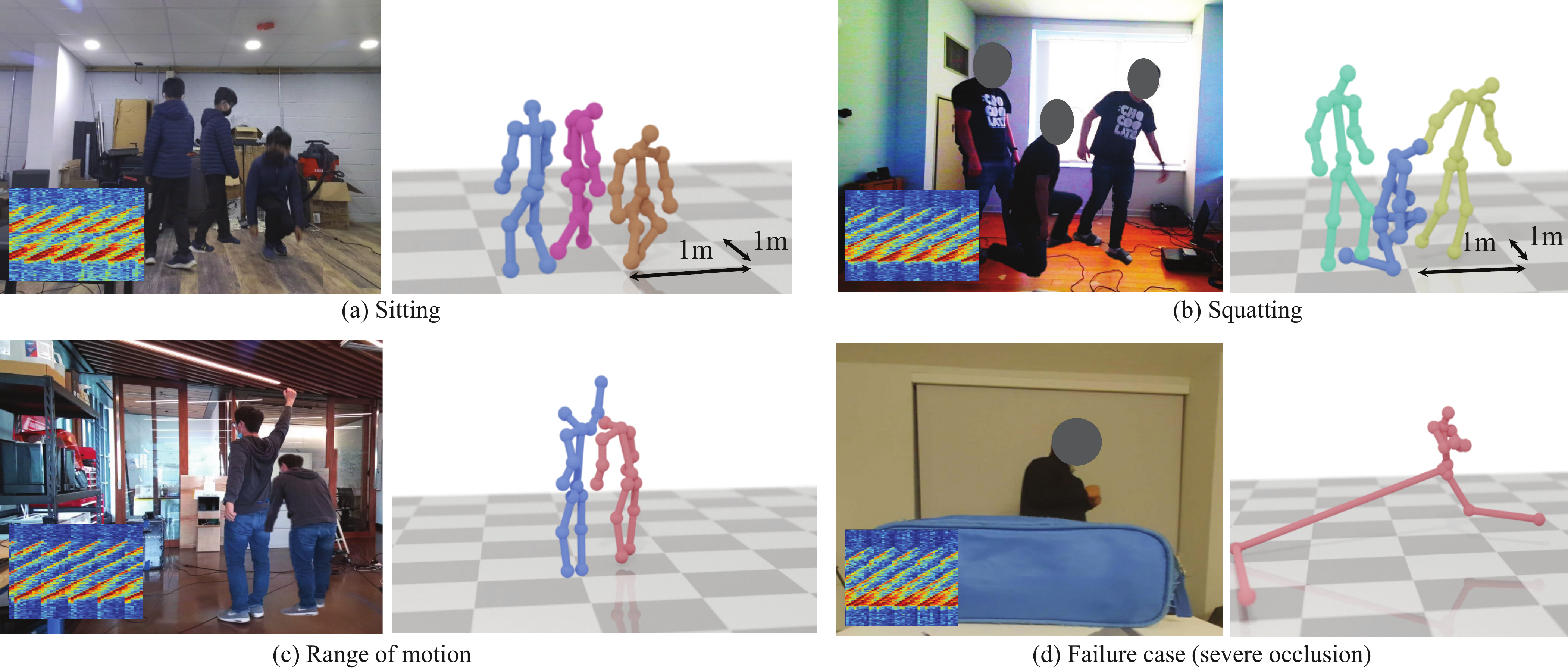}
\caption{Qualitative results. We test our pose kernel lifting approach in diverse environments including (a) basement, (b) living room, (c) laboratory, etc. The participants are asked to perform daily activities such as sitting, squatting, and range of motion. (d): A failure case of our method: severe occlusion.}
\label{fig:result_vis} 
 \vspace{-0.05in}
\end{figure*}
 \vspace{-0.05in}
 
Among the six environments in PoseKernel dataset, we use 4 environments for training and 2 environments for testing. The training data consists of diverse poses performed by six 
adult (whose heights range between 155 cm and 180 cm) and two minors (with heights 140 cm and 150 cm). The testing data includes two adult and one minor participants whose heights range between 140 cm and 180 cm. 


\noindent\textbf{Comparison} We measure the reconstruction accuracy using MPJPE metric summarized in Table~\ref{mpjpe}. As expected, state-of-the-art vision based lifting approaches (\texttt{Vis.:LfD} and \texttt{Vis.:Frank}) that predict 3D human pose in a scale-free space are sensitive to the heights of the subjects, resulting in 18 $\sim$ 40 cm mean error for adults and 40 $\sim$ 60 cm for minors, i.e., the error is larger for minor participants because their heights are very different from the average height 1.7 m. \texttt{Vis.:Frank} outperforms \texttt{Vis.:LfD}, because \texttt{Vis.:Frank} uses a larger training data, and thus estimates poses more accurately. Nonetheless, our pose kernel that is designed for metric scale reconstruction significantly outperforms these approaches. The performance is not dependent on the heights of the participants. In fact, it produces around 20\% smaller error for minor participants that the adult participants because of their smaller scale. Similar observations can be made in PCK summarized in Table~\ref{pck}. 

\noindent\textbf{Ablation Study} We ablate the sensing components of our pose kernel approach. As summarized in Tables~\ref{mpjpe} and \ref{pck}, the 3D metric lifting leverage the strong cue from visual data. Without visual cue, i.e., $\texttt{Audio$\times$4}$, the reconstruction is highly erroneous, while combining with audios as a complementary signal (\texttt{Vis.+Audio$\times$2} and \texttt{Ours}) significantly improve the accuracy. 
While providing metric information, reconstructing the 3D human pose from audio signals alone (\texttt{Audio$\times$4}) is very challenging because the signals are (1) non-directional: a received signal is integration of audio signals over all angles around the microphone which does not provide bearing angle unlike visual data; (2) non-identifiable: the reflected audio signals are not associated with any semantic information, e.g., hand, arm, and head, so it is difficult to tell where a specific reflection is coming from; and (3) slow: due to the requirement of linear frequency sweeping (10 Hz), the received signals are blurred in the presence of body motion, which is equivalent to an extremely blurry image created by 100 ms exposure with a rolling shutter. Nonetheless, augmenting audio signals improve 3D metric reconstruction regardless the heights of the participants. 

\noindent\textbf{Generalization} We report the results in completely different testing environments, which show the strong generalization ability of our method. For each environment, the spatial arrangement of the camera and audio/speakers is different, depending on the space configuration. Figure \ref{fig:result_vis} visualizes the qualitative results of our 3D pose lifting method where we successfully recover the metric scale 3D pose in different environments. We also include a failure cases in the presence of severe occlusion as shown in Figure \ref{fig:result_vis}(d).

%% file: limitations.tex
\section{Summary and Discussion} \label{sec:discussion}

This paper presented a new method to reconstruct 3D human body pose with the metric scale from a single image by leveraging audio signals. We hypothesized that the audio signals that traverse a 3D space are transformed by the human body pose through reflection, which allows us to recover the 3D metric scale pose. In order to prove this hypothesis, we use a human impulse response called pose kernel that can be spatially encoded in 3D. With the spatial encoding of the pose kernel, we learned a 3D convolutional neural network that can fuse the 2D pose detection from an image with the pose kernels to reconstruct 3D metric scale pose. We showed that our method is highly generalizable, agnostic to the room geometry, spatial arrangement of camera and speakers/microphones, and audio source signals. 


The main assumption of the pose kernel is that the room is large enough to minimize its shadow effect: in theory, there exist room impulse responses that can be canceled by the pose because the human body can occlude the room impulse response behind the person. This shadow effect is a function of room geometry, and therefore, it is dependent on the spatial arrangement of camera and speakers. In practice, we use a room, or open space larger than 5 m$\times$5 m where the impact of shadow can be neglected.



%% file: main.bbl
\begin{thebibliography}{10}\itemsep=-1pt

\bibitem{akilan2020multimodality}
Thangarajah Akilan, Edna Johnson, Gaurav Taluja, Japneet Sandhu, and Ritika
  Chadha.
\newblock Multimodality weight and score fusion for slam.
\newblock In {\em IEEE Canadian Conference on Electrical and Computer
  Engineering}, 2020.

\bibitem{arnab19}
Anurag Arnab, Carl Doersch, and Andrew Zisserman.
\newblock Exploiting temporal context for 3d human pose estimation in the wild.
\newblock In {\em CVPR}, 2019.

\bibitem{burnsmulti}
Alexis Burns, Xiaoran Fan, Jade Pinkenburg, Daewon Lee, Volkan Isler, and
  Daniel Lee.
\newblock Multi-modal dataset for human grasping.
\newblock In {\em The 29th International Conference on Robot and Human
  Interactive Communication Workshop}, 2020.

\bibitem{cai19}
Yujun Cai, Liuhao Ge, Jun Liu, Jianfei Cai, Tat-Jen Cham, Junsong Yuan, and
  Nadia~Magnenat Thalmann.
\newblock Exploiting spatial-temporal relationships for 3d pose estimation via
  graph convolutional networks.
\newblock In {\em ICCV}, 2019.

\bibitem{8765346}
Z. {Cao}, G. {Hidalgo Martinez}, T. {Simon}, S. {Wei}, and Y.~A. {Sheikh}.
\newblock Openpose: Realtime multi-person 2d pose estimation using part
  affinity fields.
\newblock {\em TPAMI}, 2019.

\bibitem{chang}
Ju~Yong Chang, Gyeongsik Moon, and Kyoung~Mu Lee.
\newblock Poselifter: Absolute 3d human pose lifting network from a single
  noisy 2d human pose.
\newblock {\em arXiv}, 2019.

\bibitem{chen2020soundspaces}
Changan Chen, Unnat Jain, Carl Schissler, Sebastia Vicenc~Amengual Gari, Ziad
  Al-Halah, Vamsi~Krishna Ithapu, Philip Robinson, and Kristen Grauman.
\newblock Soundspaces: Audio-visual navigation in 3d environments.
\newblock In {\em ECCV}, 2020.

\bibitem{chen}
Ching-Hang Chen, Ambrish Tyagi, Amit Agrawal, Dylan Drover, Stefan Stojanov,
  and James~M Rehg.
\newblock Unsupervised 3d pose estimation with geometric self-supervision.
\newblock In {\em CVPR}, 2019.

\bibitem{yu20}
Yu Cheng, Bo Yang, Bo Wang, and Robby~T Tan.
\newblock 3d human pose estimation using spatio-temporal networks with explicit
  occlusion training.
\newblock In {\em AAAI}, 2020.

\bibitem{christensen2020batvision}
Jesper~Haahr Christensen, Sascha Hornauer, and X~Yu Stella.
\newblock Batvision: Learning to see 3d spatial layout with two ears.
\newblock In {\em ICRA}, 2020.

\bibitem{ci19}
Hai Ci, Chunyu Wang, Xiaoxuan Ma, and Yizhou Wang.
\newblock Optimizing network structure for 3d human pose estimation.
\newblock In {\em ICCV}, 2019.

\bibitem{doherty2019multimodal}
Kevin Doherty, Dehann Fourie, and John Leonard.
\newblock Multimodal semantic slam with probabilistic data association.
\newblock In {\em ICRA}, 2019.

\bibitem{ephrat:2018}
Ariel Ephrat, Inbar Mosseri, Oran Lang, Tali Dekel, Kevin Wilson, Avinatan
  Hassidim, William~T. Freeman, and Michael Rubinstein.
\newblock Looking to listen at the cocktail party: A speaker-independent
  audio-visual model for speech separation.
\newblock In {\em arXiv}, 2018.

\bibitem{fan2020acoustic}
Xiaoran Fan, Daewon Lee, Yuan Chen, Colin Prepscius, Volkan Isler, Larry
  Jackel, H~Sebastian Seung, and Daniel Lee.
\newblock Acoustic collision detection and localization for robot manipulators.
\newblock In {\em IROS}, 2020.

\bibitem{fan2021aurasense}
Xiaoran Fan, Riley Simmons-Edler, Daewon Lee, Larry Jackel, Richard Howard, and
  Daniel Lee.
\newblock Aurasense: Robot collision avoidance by full surface proximity
  detection.
\newblock {\em arXiv}, 2021.

\bibitem{gao:2019}
R. Gao and K. Grauman.
\newblock Co-separating sounds of visual objects.
\newblock In {\em ICCV}, 2019.

\bibitem{ghaleb2019metric}
Esam Ghaleb, Mirela Popa, and Stylianos Asteriadis.
\newblock Metric learning-based multimodal audio-visual emotion recognition.
\newblock {\em IEEE Multimedia}, 2019.

\bibitem{gong}
Kehong Gong, Jianfeng Zhang, and Jiashi Feng.
\newblock Poseaug: A differentiable pose augmentation framework for 3d human
  pose estimation.
\newblock In {\em CVPR}, 2021.

\bibitem{guan2020through}
Junfeng Guan, Sohrab Madani, Suraj Jog, Saurabh Gupta, and Haitham Hassanieh.
\newblock Through fog high-resolution imaging using millimeter wave radar.
\newblock In {\em CVPR}, 2020.

\bibitem{habibie}
Ikhsanul Habibie, Weipeng Xu, Dushyant Mehta, Gerard Pons-Moll, and Christian
  Theobalt.
\newblock In the wild human pose estimation using explicit 2d features and
  intermediate 3d representations.
\newblock In {\em CVPR}, 2019.

\bibitem{he20}
Yihui He, Rui Yan, Katerina Fragkiadaki, and Shoou-I Yu.
\newblock Epipolar transformers.
\newblock In {\em CVPR}, 2020.

\bibitem{ionescu2013human36m}
Catalin Ionescu, Dragos Papava, Vlad Olaru, and Cristian Sminchisescu.
\newblock Human3.6m: Large scale datasets and predictive methods for 3d human
  sensing in natural environments.
\newblock {\em TPAMI}, 2013.

\bibitem{iqbal}
Umar Iqbal, Pavlo Molchanov, and Jan Kautz.
\newblock Weakly-supervised 3d human pose learning via multi-view images in the
  wild.
\newblock In {\em CVPR}, 2020.

\bibitem{iskakov}
Karim Iskakov, Egor Burkov, Victor Lempitsky, and Yury Malkov.
\newblock Learnable triangulation of human pose.
\newblock In {\em ICCV}, 2019.

\bibitem{wipose}
Wenjun Jiang, Hongfei Xue, Chenglin Miao, Shiyang Wang, Sen Lin, Chong Tian,
  Srinivasan Murali, Haochen Hu, Zhi Sun, and Lu Su.
\newblock Towards 3d human pose construction using wifi.
\newblock In {\em Annual International Conference on Mobile Computing and
  Networking}, 2020.

\bibitem{jin2018towards}
Haojian Jin, Zhijian Yang, Swarun Kumar, and Jason~I Hong.
\newblock Towards wearable everyday body-frame tracking using passive rfids.
\newblock {\em ACM Interactive, Mobile, Wearable and Ubiquitous Technologies},
  2018.

\bibitem{joo2020eft}
Hanbyul Joo, Natalia Neverova, and Andrea Vedaldi.
\newblock Exemplar fine-tuning for 3d human pose fitting towards in-the-wild 3d
  human pose estimation.
\newblock {\em 3DV}, 2021.

\bibitem{kocabas}
Muhammed Kocabas, Salih Karagoz, and Emre Akbas.
\newblock Self-supervised learning of 3d human pose using multi-view geometry.
\newblock In {\em CVPR}, 2019.

\bibitem{kudo}
Yasunori Kudo, Keisuke Ogaki, Yusuke Matsui, and Yuri Odagiri.
\newblock Unsupervised adversarial learning of 3d human pose from 2d joint
  locations.
\newblock {\em arXiv}, 2018.

\bibitem{li_cvpr21}
Jiefeng Li, Chao Xu, Zhicun Chen, Siyuan Bian, Lixin Yang, and Cewu Lu.
\newblock Hybrik: A hybrid analytical-neural inverse kinematics solution for 3d
  human pose and shape estimation.
\newblock In {\em CVPR}, 2021.

\bibitem{li21}
Wenhao Li, Hong Liu, Runwei Ding, Mengyuan Liu, Pichao Wang, and Wenming Yang.
\newblock Exploiting temporal contexts with strided transformer for 3d human
  pose estimation.
\newblock {\em arXiv}, 2021.

\bibitem{liu2018towards}
Hongyi Liu, Tongtong Fang, Tianyu Zhou, and Lihui Wang.
\newblock Towards robust human-robot collaborative manufacturing: Multimodal
  fusion.
\newblock {\em IEEE Access}, 2018.

\bibitem{liu20}
Ruixu Liu, Ju Shen, He Wang, Chen Chen, Sen-ching Cheung, and Vijayan Asari.
\newblock Attention mechanism exploits temporal contexts: Real-time 3d human
  pose reconstruction.
\newblock In {\em CVPR}, 2020.

\bibitem{llopart}
Adrian Llopart.
\newblock Liftformer: 3d human pose estimation using attention models.
\newblock {\em arXiv}, 2020.

\bibitem{SMPL:2015}
Matthew Loper, Naureen Mahmood, Javier Romero, Gerard Pons-Moll, and Michael~J.
  Black.
\newblock {SMPL}: A skinned multi-person linear model.
\newblock {\em TOG}, 2015.

\bibitem{mono-3dhp2017}
Dushyant Mehta, Helge Rhodin, Dan Casas, Pascal Fua, Oleksandr Sotnychenko,
  Weipeng Xu, and Christian Theobalt.
\newblock Monocular 3d human pose estimation in the wild using improved cnn
  supervision.
\newblock In {\em 3DV}, 2017.

\bibitem{xneck}
Dushyant Mehta, Oleksandr Sotnychenko, Franziska Mueller, Weipeng Xu, Mohamed
  Elgharib, Pascal Fua, Hans-Peter Seidel, Helge Rhodin, Gerard Pons-Moll, and
  Christian Theobalt.
\newblock Xnect: Real-time multi-person 3d motion capture with a single rgb
  camera.
\newblock {\em TOG}, 2020.

\bibitem{vneck}
Dushyant Mehta, Srinath Sridhar, Oleksandr Sotnychenko, Helge Rhodin, Mohammad
  Shafiei, Hans-Peter Seidel, Weipeng Xu, Dan Casas, and Christian Theobalt.
\newblock Vnect: Real-time 3d human pose estimation with a single rgb camera.
\newblock {\em TOG}, 2017.

\bibitem{mroueh2015deep}
Youssef Mroueh, Etienne Marcheret, and Vaibhava Goel.
\newblock Deep multimodal learning for audio-visual speech recognition.
\newblock In {\em ICASSP}, 2015.

\bibitem{nadon2018multi}
F{\'e}lix Nadon, Angel~J Valencia, and Pierre Payeur.
\newblock Multi-modal sensing and robotic manipulation of non-rigid objects: A
  survey.
\newblock {\em Robotics}, 2018.

\bibitem{ngiam2011multimodal}
Jiquan Ngiam, Aditya Khosla, Mingyu Kim, Juhan Nam, Honglak Lee, and Andrew~Y
  Ng.
\newblock Multimodal deep learning.
\newblock In {\em ICML}, 2011.

\bibitem{owens2018audio}
Andrew Owens and Alexei~A Efros.
\newblock Audio-visual scene analysis with self-supervised multisensory
  features.
\newblock {\em ECCV}, 2018.

\bibitem{pavlakos}
Georgios Pavlakos, Xiaowei Zhou, Konstantinos~G Derpanis, and Kostas
  Daniilidis.
\newblock Coarse-to-fine volumetric prediction for single-image 3d human pose.
\newblock In {\em CVPR}, 2017.

\bibitem{pavllo}
Dario Pavllo, Christoph Feichtenhofer, David Grangier, and Michael Auli.
\newblock 3d human pose estimation in video with temporal convolutions and
  semi-supervised training.
\newblock In {\em CVPR}, 2019.

\bibitem{hossain}
Mir Rayat Imtiaz~Hossain and James~J Little.
\newblock Exploiting temporal information for 3d pose estimation.
\newblock {\em arXiv}, 2017.

\bibitem{rhodin_eccv18}
Helge Rhodin, Mathieu Salzmann, and Pascal Fua.
\newblock Unsupervised geometry-aware representation for 3d human pose
  estimation.
\newblock In {\em ECCV}, 2018.

\bibitem{rhodin18}
Helge Rhodin, J{\"o}rg Sp{\"o}rri, Isinsu Katircioglu, Victor Constantin,
  Fr{\'e}d{\'e}ric Meyer, Erich M{\"u}ller, Mathieu Salzmann, and Pascal Fua.
\newblock Learning monocular 3d human pose estimation from multi-view images.
\newblock In {\em CVPR}, 2018.

\bibitem{rodriguez2018methodology}
Saith Rodr{\'\i}guez, Carlos~A Quintero, Andrea~K P{\'e}rez, Eyberth Rojas,
  Oswaldo Pe{\~n}a, and Fernando De~La~Rosa.
\newblock Methodology for learning multimodal instructions in the context of
  human-robot interaction using machine learning.
\newblock In {\em International Symposium on Intelligent Computing Systems},
  2018.

\bibitem{rong2021frankmocap}
Yu Rong, Takaaki Shiratori, and Hanbyul Joo.
\newblock Frankmocap: A monocular 3d whole-body pose estimation system via
  regression and integration.
\newblock In {\em ICCV Workshops}, 2021.

\bibitem{schoenberger2016sfm}
Johannes~Lutz Sch\"{o}nberger and Jan-Michael Frahm.
\newblock Structure-from-motion revisited.
\newblock In {\em CVPR}, 2016.

\bibitem{sengupta2019dnn}
Arindam Sengupta, Feng Jin, and Siyang Cao.
\newblock A dnn-lstm based target tracking approach using mmwave radar and
  camera sensor fusion.
\newblock In {\em IEEE National Aerospace and Electronics Conference}, 2019.

\bibitem{senocak2018learning}
Arda Senocak, Tae-Hyun Oh, Junsik Kim, Ming-Hsuan Yang, and In~So Kweon.
\newblock Learning to localize sound source in visual scenes.
\newblock In {\em CVPR}, 2018.

\bibitem{shen2016smartwatch}
Sheng Shen, He Wang, and Romit Roy~Choudhury.
\newblock I am a smartwatch and i can track my user's arm.
\newblock In {\em Annual International Conference on Mobile Systems,
  Applications, and Services}, 2016.

\bibitem{singhal2016multi}
Prateek Singhal, Ruffin White, and Henrik Christensen.
\newblock Multi-modal tracking for object based slam.
\newblock {\em arXiv}, 2016.

\bibitem{sun2018}
Xiao Sun, Bin Xiao, Fangyin Wei, Shuang Liang, and Yichen Wei.
\newblock Integral human pose regression.
\newblock In {\em ECCV}, 2018.

\bibitem{cjtaylor}
Camillo~J Taylor.
\newblock Reconstruction of articulated objects from point correspondences in a
  single uncalibrated image.
\newblock {\em Computer Vision and Image Understanding}, 2000.

\bibitem{terblanche2021multimodal}
Johan Terblanche, Sam Claassens, and Dehann Fourie.
\newblock Multimodal navigation-affordance matching for slam.
\newblock {\em IEEE Robotics and Automation Letters}, 2021.

\bibitem{tian2021cyclic}
Yapeng Tian, Di Hu, and Chenliang Xu.
\newblock Cyclic co-learning of sounding object visual grounding and sound
  separation.
\newblock In {\em CVPR}, 2021.

\bibitem{tome}
Denis Tome, Chris Russell, and Lourdes Agapito.
\newblock Lifting from the deep: Convolutional 3d pose estimation from a single
  image.
\newblock In {\em CVPR}, 2017.

\bibitem{Tripathi20}
Shashank Tripathi, Siddhant Ranade, Ambrish Tyagi, and Amit Agrawal.
\newblock Posenet3d: Learning temporally consistent 3d human pose via knowledge
  distillation.
\newblock In {\em 3DV}, 2020.

\bibitem{varol17_surreal}
G{\"u}l Varol, Javier Romero, Xavier Martin, Naureen Mahmood, Michael~J. Black,
  Ivan Laptev, and Cordelia Schmid.
\newblock Learning from synthetic humans.
\newblock In {\em CVPR}, 2017.

\bibitem{vonMarcard2018}
Timo von Marcard, Roberto Henschel, Michael Black, Bodo Rosenhahn, and Gerard
  Pons-Moll.
\newblock Recovering accurate 3d human pose in the wild using imus and a moving
  camera.
\newblock In {\em ECCV}, 2018.

\bibitem{wandt19}
Bastian Wandt and Bodo Rosenhahn.
\newblock Repnet: Weakly supervised training of an adversarial reprojection
  network for 3d human pose estimation.
\newblock In {\em CVPR}, 2019.

\bibitem{wendt}
Bastian Wandt, Marco Rudolph, Petrissa Zell, Helge Rhodin, and Bodo Rosenhahn.
\newblock Canonpose: Self-supervised monocular 3d human pose estimation in the
  wild.
\newblock In {\em CVPR}, 2021.

\bibitem{wang2019multimodal}
Tao Wang, Chao Yang, Frank Kirchner, Peng Du, Fuchun Sun, and Bin Fang.
\newblock Multimodal grasp data set: A novel visual--tactile data set for
  robotic manipulation.
\newblock {\em International Journal of Advanced Robotic Systems}, 2019.

\bibitem{wang2016robot}
Zhichao Wang, Zhiqi Li, Bin Wang, and Hong Liu.
\newblock Robot grasp detection using multimodal deep convolutional neural
  networks.
\newblock {\em Advances in Mechanical Engineering}, 2016.

\bibitem{wang20}
Zhe Wang, Daeyun Shin, and Charless~C Fowlkes.
\newblock Predicting camera viewpoint improves cross-dataset generalization for
  3d human pose estimation.
\newblock In {\em ECCV}, 2020.

\bibitem{watkins2019multi}
David Watkins-Valls, Jacob Varley, and Peter Allen.
\newblock Multi-modal geometric learning for grasping and manipulation.
\newblock In {\em ICRA}, 2019.

\bibitem{wei2016cpm}
Shih-En Wei, Varun Ramakrishna, Takeo Kanade, and Yaser Sheikh.
\newblock Convolutional pose machines.
\newblock In {\em CVPR}, 2016.

\bibitem{wilson2021echo}
Justin Wilson, Nicholas Rewkowski, Ming~C Lin, and Henry Fuchs.
\newblock Echo-reconstruction: Audio-augmented 3d scene reconstruction.
\newblock {\em arXiv}, 2021.

\bibitem{xu21}
Tianhan Xu and Wataru Takano.
\newblock Graph stacked hourglass networks for 3d human pose estimation.
\newblock In {\em CVPR}, 2021.

\bibitem{yang2020ear}
Zhijian Yang, Yu-Lin Wei, Sheng Shen, and Romit~Roy Choudhury.
\newblock Ear-ar: indoor acoustic augmented reality on earphones.
\newblock In {\em Annual International Conference on Mobile Computing and
  Networking}, 2020.

\bibitem{yao19}
Yuan Yao, Yasamin Jafarian, and Hyun~Soo Park.
\newblock Monet: Multiview semi-supervised keypoint detection via epipolar
  divergence.
\newblock In {\em ICCV}, 2019.

\bibitem{yushuke}
Yusuke Yoshiyasu, Ryusuke Sagawa, Ko Ayusawa, and Akihiko Murai.
\newblock Skeleton transformer networks: 3d human pose and skinned mesh from
  single rgb image.
\newblock In {\em ACCV}, 2018.

\bibitem{yun2015turning}
Sangki Yun, Yi-Chao Chen, and Lili Qiu.
\newblock Turning a mobile device into a mouse in the air.
\newblock In {\em Annual International Conference on Mobile Systems,
  Applications, and Services}, 2015.

\bibitem{zhao19}
Long Zhao, Xi Peng, Yu Tian, Mubbasir Kapadia, and Dimitris~N Metaxas.
\newblock Semantic graph convolutional networks for 3d human pose regression.
\newblock In {\em CVPR}, 2019.

\bibitem{rfpose}
Mingmin Zhao, Tianhong Li, Mohammad Abu~Alsheikh, Yonglong Tian, Hang Zhao,
  Antonio Torralba, and Dina Katabi.
\newblock Through-wall human pose estimation using radio signals.
\newblock In {\em CVPR}, 2018.

\bibitem{rfavatar}
Mingmin Zhao, Yingcheng Liu, Aniruddh Raghu, Tianhong Li, Hang Zhao, Antonio
  Torralba, and Dina Katabi.
\newblock Through-wall human mesh recovery using radio signals.
\newblock In {\em ICCV}, 2019.

\end{thebibliography}
